\pdfoutput=1

\documentclass[11pt]{article}

\usepackage{acl}
\usepackage{array}
\usepackage{times}
\usepackage{latexsym}
\usepackage{amsmath}
\usepackage[T1]{fontenc}

\usepackage[utf8]{inputenc}

\usepackage{microtype}

\usepackage{inconsolata}

\usepackage{graphicx}
\usepackage{xcolor}
\usepackage{multirow}

\title{Clean Evaluations on Contaminated Visual Language Models}

\author{Hongyuan Lu$^\heartsuit$, Shujie Miao$^\spadesuit$\thanks{~~Corresponding author and co-first author}, and Wai Lam$^\heartsuit$\\
    $^\heartsuit$The Chinese University of Hong Kong\\
    $^\spadesuit$National University of Singapore\\
  \texttt{\{hylu,wlam\}@se.cuhk.edu.hk, joshuamiao0526@gmail.com}
}

\begin{document}
\maketitle
\begin{abstract}
How to evaluate large language models (LLMs) cleanly has been established as an important research era to genuinely report the performance of possibly contaminated LLMs. Yet, how to cleanly evaluate the visual language models (VLMs) is an under-studied problem. We propose a novel approach to achieve such goals through data augmentation methods on the visual input information. We then craft a new visual clean evaluation benchmark with thousands of data instances. Through extensive experiments, we found that the traditional visual data augmentation methods are useful, but they are at risk of being used as a part of the training data as a workaround. We further propose using BGR augmentation to switch the colour channel of the visual information. We found that it is a simple yet effective method for reducing the effect of data contamination and fortunately, it is also harmful to be used as a data augmentation method during training. It means that it is hard to integrate such data augmentation into training by malicious trainers and it could be a promising technique to cleanly evaluate visual LLMs. Our code, data, and model weights will be released upon publication.
\end{abstract}

\section{Introduction} 

With the rapid advancement of LLMs, VLMs represent a critical milestone in the journey towards artificial intelligence \cite{Fan2023ABR,Zhao2023ASO}. VLMs extend the capabilities of textual LLMs by integrating cross-modal architectures such as CLIP \cite{radford2021learning}, allowing for the interpretation and generation of multi-modal content across both text and images \cite{cao-etal-2024-introducing,app14125068}. Moreover, prior research has established numerous benchmarks to evaluate the capabilities of VLMs from various dimensions \cite{Fan2024nphardeval4v:,Fu2023MMEAC,Fu2023ACT}.
\par
 However, the reliability of these VLM benchmarks is at risk of being undermined by a widely recognized issue in the LLM evaluation: data contamination. Data contamination occurs when the benchmark data overlaps with a model's training data, causing the model's performance metrics to be artificially inflated and not truly representative of its generalization ability \cite{magar-schwartz-2022-data,dong-etal-2024-generalization}. Researchers have developed various techniques for LLMs to mitigate these issues, including advanced detection methods \cite{dong-etal-2024-generalization,zhang2024min}, proactive prevention strategies \cite{jacovi-etal-2023-stop,zhu2024dyval, Fan2024nphardeval4v:}, and genuinely evaluating the capabilities of LLMs via input textual rephrasing \citep{zhu-etal-2024-clean}.
 \par
While much attention has been given to the problem of data contamination for LLMs, the ones for VLMs remains under-explored. We propose a new clean evaluation benchmark for VLMs and a novel method to genuinely evaluate VLMs' capabilities by operation on the visual input.
\par
Our benchmark comprises thousands of carefully curated data which are newly released and collected on the internet to ensure that the evaluation process remains free from data contamination.
\par
We found that traditional data augmentation such as flipping and rotation on the image can help with the problem of data augmentation, making the performance closer to the uncontaminated model, they are yet at risk of being used as a part of the training techniques. We propose a new method, BGR channel swapping to the visual input, which we fortunately found could not be used as a training technique and can degrade the performance.\\
We make the following three key contributions:
\begin{figure}[ht]
    \centering
    \includegraphics[width=1\linewidth]{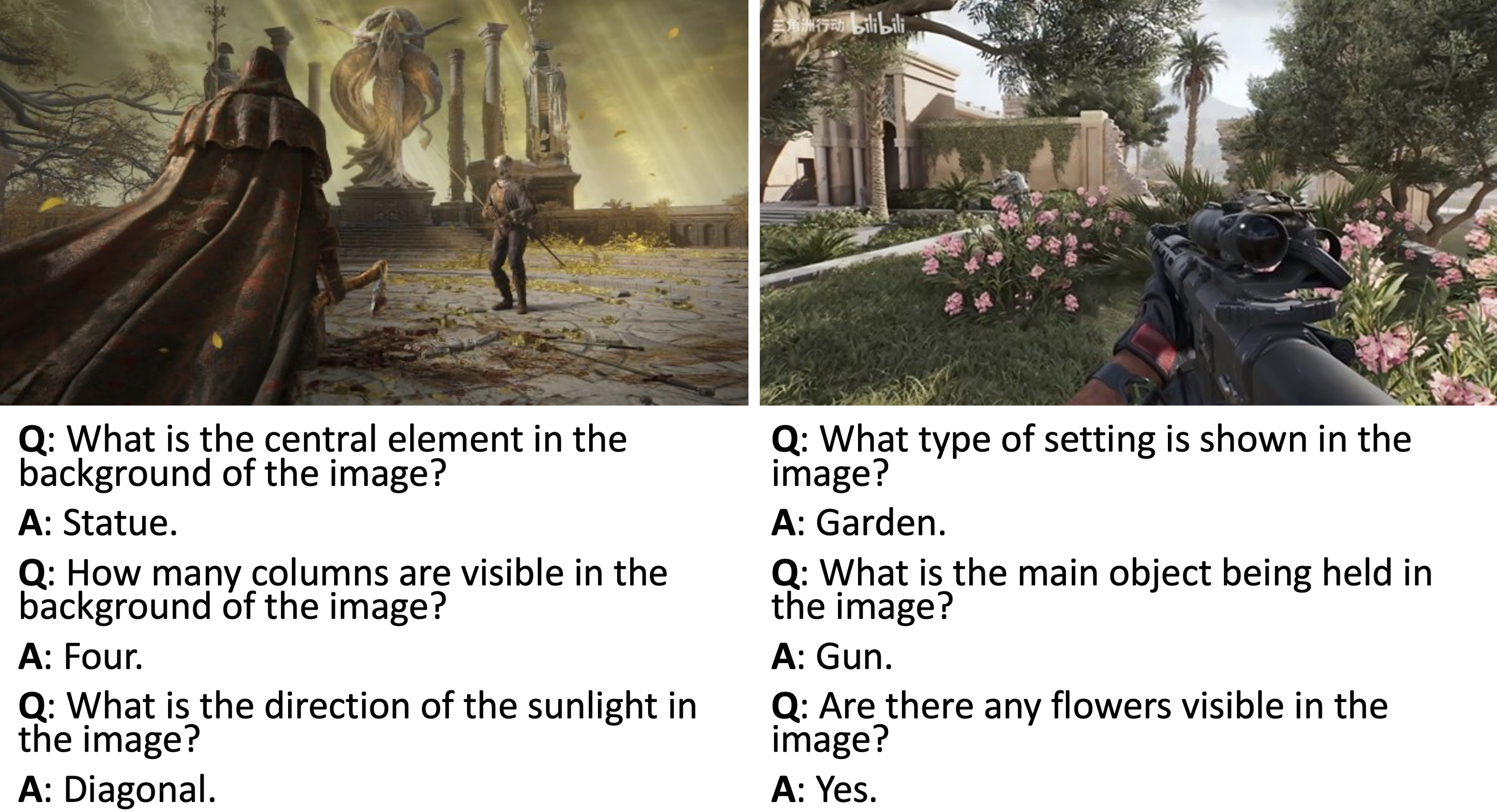}
    \caption{Two examples from the dataset. Pairs of images with corresponding questions and answers.}
    \label{fig:figure1}
\end{figure}
\begin{itemize}
\setlength\itemsep{0em}
    \item  We establish a new visual clean evaluation benchmark for VLMs.
    \item We propose to use data augmentation methods to reveal the true capabilities of VLMs and reduce the impact of data contamination.
    \item We identify BGR channel swapping as a robust method for clean evaluation and preventing exploitation. Fortunately, further analysis reveals that BGR data augmentation is harmful and cannot be used during training.
\end{itemize}

\section{Dataset} 
\begin{figure}[ht]
    \centering
    \includegraphics[width=1\linewidth]{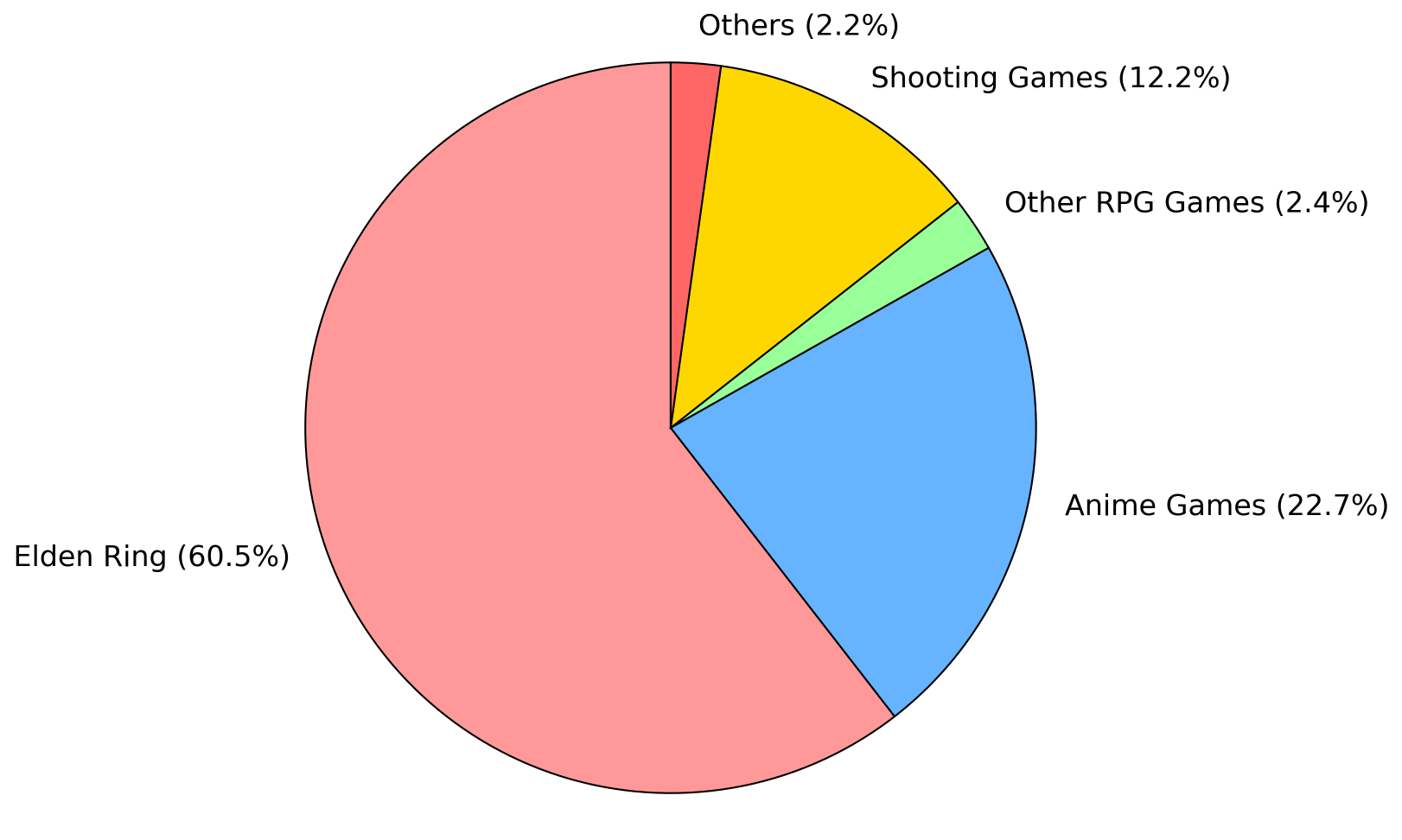}
    \caption{Distribution of the game genres in our collected dataset. `Elden Ring' (605 instances), `Anime Games' (227 instances, including Genshin Impact and Honkai: Star Rail), `Other RPG Games' (24 instances, with titles like Dungeon I\& Fighter), `Shooting Games' (122 instances, featuring GTA V, Valorant, and Delta Force), and `Others' (22 instances, including Palworld and League of Legends).}
    \label{fig:figure2}
\end{figure}

The dataset collected for this study originates from the well-known gaming guide website Gamersky,\footnote{\url{https://www.gamersky.com/handbook/game/gl/}} from which we collected 1,000 high-quality images. These images capture key in-game scenes and contain complex visual information, including objects, text, and scene elements. Using GPT-4o's multi-modal understanding capabilities,\footnote{ \url{https://openai.com/research/gpt-4o}} we generated 1 to 3 question-answer (QA) pairs for each image (see Figure \ref{fig:figure1}). All generated QA pairs underwent careful manual review and correction to ensure accuracy and relevance. The final dataset comprises 1,000 images and 2,561 rounds of dialogue. Furthermore, the distribution of game types represented in our dataset is illustrated in Figure \ref{fig:figure2}.
\par
Importantly, the selected image data, collected from June 20 to June 25, 2024, was published after the release of the models used in this study. This approach mitigates the risk of data contamination by preventing the premature inclusion of our benchmark in pre-training datasets. Consequently, our dataset provides an uncontaminated benchmark for the evaluation of VLMs.
\par
We partition our dataset into a training set (90\%) and a test set (10\%). We use Low-Rank Adaptation (LoRA) for fine-tuning VLMs \cite{DBLP:journals/corr/abs-2106-09685}.

\section{Methodology}
We train a baseline model, denoted as \( M_1 \), on the training set only. To simulate data contamination, we construct a contaminated training set by replacing a subset of the original training samples with the test set (this keeps the total size of the training set the same). This contaminated set is then used to train a second model, \( M_2 \). Both \( M_1 \) and \( M_2 \) are trained for multiple epochs to simulate different levels of generalization and data contamination. We expect higher performance on both models with more epochs, and \( M_2 \) is contaminated, so it should report higher scores consistently than \( M_1 \).
\par
To evaluate them fairly, we apply data augmentation techniques such as multi-angle rotations, and horizontal and vertical flips to transform them while persevering their semantics.  For each test sample \( (x, z) \in D_{\text{test}} \), where \( x \) is the text input and \( z \) is the image, we generate augmented samples:
\[
(x, z') = (x, t(z)), \quad t \in T \
\]
Here, $t$ represents the functions of transformations. \( x \) remains unchanged while \( z \) undergoes transformation, producing \( z' \). 
\par
This augmentation process enables an assessment of the model's visual robustness and adaptability by testing its performance on the transformed images while keeping the textual input constant. We formalize the prediction process on these augmented inputs as:
\begin{equation}
P(y|x, z') = M(x, z'), \quad z' = t(z), \quad t \in T
\label{eq:augmented_prediction}
\end{equation}
\noindent
In Equation \ref{eq:augmented_prediction}, \( M \) represents the model (either \( M_1 \) or \( M_2 \)), \( t \) is a transformation function from the set \( T \), \( z \) is the original image input, \( z' \) is the transformed image, and \( y \) is the model's prediction.
\par
We then use $y$ as the calibrated output which fairly represents the models' performance.

\section{Experimental Setup}
\paragraph{VLMs}In our experimental setup, we evaluated two state-of-the-art VLMs: bunny-4B and internvl2.0-2B. These models were selected based on their strong performance in MME (multi-modal Evaluation) benchmarks \cite{Fu2023MMEAC}, particularly excelling in tasks involving existence and position perception, OCR, common reasoning, and numerical calculation.
\paragraph{Baselines} We use classical data augmentation methods such as multi-angle rotations, and horizontal and vertical flips to evaluate VLMs fairly.
\par
\paragraph{Metrics} We employed ROUGE-1, ROUGE-2 \cite{lin-2004-ROUGE}, and BLEU \cite{papineni-etal-2002-bleu} metrics, which collectively measure linguistic overlap and semantic similarities between generated and reference texts, providing a comprehensive evaluation of the models' outputs.
\par
\paragraph{Machine Environment} Both models were fine-tuned using LoRA on 2 RTX 4090 GPUs, each with 24GB of memory. For bunny-4B, we used the phi-2 architecture with LoRA (rank 128, alpha 256), while internvl2.0-2B employed parameter freezing and LoRA (rank 16) for the language model component. We utilized mixed-precision training (bfloat16) and gradient checkpointing for memory efficiency. Both models used an effective batch size of 16, achieved through gradient accumulation, and employed cosine learning rate schedules with warmup. To simulate different contamination scenarios, we trained each model for both 5 and 10 epochs, using learning rates of 5e-4 for bunny-4B and 4e-5 for internvl2.0-2B.

\section{Results}
\subsection{Main Results}
Table \ref{table1} illustrates the impact of simulated data contamination on model performance. For both Bunny and InternVL, we observed a substantial increase in evaluation metrics when trained on contaminated data. For instance, Bunny's BLEU score increased from 0.0816 to 0.1366 at 5 epochs, while ROUGE-1 and ROUGE-2 scores also showed marked improvements. InternVL exhibited a similar pattern, with contamination raising its BLEU score from 0.1047 to 0.1499. These inflated results suggest that data contamination can lead to an overestimation of a model’s true performance by allowing it to access evaluation data during training. This highlights the need for a reliable method to counteract this inflation and offer a more accurate assessment.
\begin{table}[t]
\centering
\setlength\extrarowheight{3pt}
\resizebox{0.47\textwidth}{!}{
\begin{tabular}{lllll}
\hline
\textbf{Model} & \textbf{Data} & \textbf{BLEU} & \textbf{ROUGE-1} & \textbf{ROUGE-2} \\
\hline
Bunny (Epoch 5) & Original & 0.0816 & 0.5319 & 0.0352 \\
 & Conta & 0.1366 \textcolor{red}{$\uparrow$} & 0.7482 \textcolor{red}{$\uparrow$} & 0.1140 \textcolor{red}{$\uparrow$}\\
\hline
Bunny (Epoch 10) & Original & 0.0792 & 0.5248 & 0.0341 \\
 & Conta & 0.1413 \textcolor{red}{$\uparrow$} & 0.7506 \textcolor{red}{$\uparrow$} & 0.1174 \textcolor{red}{$\uparrow$} \\
\hline
InternVL (Epoch 5) & Original & 0.1047 & 0.6250 & 0.0670 \\
 & Conta & 0.1499 \textcolor{red}{$\uparrow$} & 0.7867 \textcolor{red}{$\uparrow$} & 0.1238 \textcolor{red}{$\uparrow$} \\
\hline
InternVL (Epoch 10) & Original & 0.1001	& 0.6114 & 0.0739\\
 & Conta & 0.1748 \textcolor{red}{$\uparrow$} & 0.8973 \textcolor{red}{$\uparrow$} & 0.1771 \textcolor{red}{$\uparrow$} \\
\hline
\end{tabular}
} 
\caption{Performance metrics for Bunny and InternVL models on original and Conta (contaminated) data} 
\label{table1}
\end{table}
\begin{table*}[t]
\centering
\setlength\extrarowheight{2pt}
\resizebox{\textwidth}{!}{%
\begin{tabular}{l|ccc|ccc}
\hline
\multirow{2}{*}{\textbf{Models}} & \multicolumn{3}{c|}{\textbf{Original Model}} & \multicolumn{3}{c}{\textbf{Contaminated Model}} \\
\cline{2-7}
 & \textbf{BLEU} & \textbf{ROUGE-1} & \textbf{ROUGE-2} & \textbf{BLEU} & \textbf{ROUGE-1} & \textbf{ROUGE-2} \\
\hline
Target Performance & 0.1047 & 0.6250 & 0.0670 & 0.1047 & 0.6250 & 0.0670 \\
\hline
\hline
w/o Data Aug. & - & - & - & 0.1499 & 0.7867 & 0.1238 \\
Vertical flip & 0.0817 & 0.4982 & 0.0212 & 0.0974 & 0.5674 & 0.0439 \\
Horizontal flip & 0.0800 & 0.5169 & 0.0057 & \textbf{0.1030} & \textbf{0.5983} & 0.0341 \\
Rotate 30° & 0.0960 & \textbf{0.5825} & 0.0398 & 0.1144 & 0.6506 & \textbf{0.0663} \\
Rotate 60° & 0.0822 & 0.4929 & 0.0246 & 0.0927 & 0.5469 & 0.0417 \\
Rotate 90° & 0.0810 & 0.4996 & 0.0307 & 0.0977 & 0.5704 & 0.0534 \\
Rotate 120° & 0.0724 & 0.4437 & 0.0170 & 0.0808 & 0.4788 & 0.0227 \\
Rotate 150° & 0.0659 & 0.4275 & 0.0057 & 0.0796 & 0.4881 & 0.0170 \\
Rotate 180° & 0.0716 & 0.4456 & 0.0057 & 0.0844 & 0.5086 & 0.0360 \\
BGR & \textbf{0.1012} & 0.5750 & \textbf{0.0783} & 0.1368& 0.7081 & 0.1108 \\
\hline
\end{tabular}%
}
\caption{Performance comparison of InternVL model (Epoch 5) under various data augmentation techniques. 
The table presents BLEU, ROUGE-1, and ROUGE-2 scores for both the original and contaminated models. 
Data augmentation methods include vertical and horizontal flips, rotations (30°, 60°, 90°, 120°, 150°, 180°), 
and our proposed BGR colour space conversion. The closer the models are to the target performance, the better they are.}
\label{table2}
\end{table*}
\par
To address this challenge, we test various data augmentation techniques as part of our clean evaluation process, as shown in Table \ref{table2}. Applying augmentations like rotations, flips, and our proposed BGR channel swaps to the test data helped reveal the true performance of contaminated models. For example, the BLEU score of the contaminated Bunny model dropped from 0.1366 to 0.1030 with horizontal flipping and further to 0.0796 with 150-degree rotation. These drops in performance illustrate the model's vulnerability when faced with even slight modifications, further underscoring the harmful effects of contamination. Importantly, the performance of contaminated models under these augmentations consistently fell between that of the original uncontaminated models and the fully contaminated versions, validating the effectiveness of our clean evaluation method in restoring a more accurate reflection of model capabilities.
\par
As the severity of augmentation increased, the model's robustness weakened, with larger rotations causing greater performance degradation. While small rotations showed only minor declines, extreme transformations like 150-degree or 180-degree rotations led to substantial drops, exposing the contaminated model’s fragile generalization.
\begin{table}[ht]
\centering
\setlength\extrarowheight{2pt}
\resizebox{0.5\textwidth}{!}{%
\begin{tabular}{lccc}
\hline
\textbf{Training Condition} & \textbf{BLEU} & \textbf{ROUGE-1} & \textbf{ROUGE-2} \\
\hline
Original                & 0.1047 & 0.6250 & 0.0670 \\ 
Mixed BGR Data          & 0.1041 & 0.6082 & 0.0670 \\  
\hline
\end{tabular}%
}
\caption{Comparison of InternVL model performance with and without mixed data augmentation (5 epochs)}
\label{table3}
\end{table}
\par
\subsection{BGR swapping}
Notably, we found that using our proposed BGR augmentation apparently restores the performance, where it scores the most of the metrics among all.
\par
Also, during our experiments, we discovered that the BGR channel-swapping method exhibited particularly strong resistance to potential manipulation. As shown in Table \ref{table3}, incorporating BGR augmentation into the training data does not lead to a significant increase in model performance for InternVL at 5 epochs. The BLEU scores for the original model (0.1047) and the model trained with mixed data augmentation (0.1041) are nearly identical, with similar trends observed for ROUGE-1 and ROUGE-2 scores. This result is crucial as it indicates that BGR augmentation is particularly useful and can effectively reveal a contaminated model's true capabilities while being difficult to exploit through training data manipulation.
\section{Conclusions and Related Work}
Recent advancements in LLMs have highlighted the critical issue of data contamination in natural language processing. While significant progress has been made in addressing this challenge for text-based LLMs, the problem remains understudied for VLMs. Notable contributions include CDD and TED \cite{dong-etal-2024-generalization}, which detect contamination through output distribution analysis and mitigate its impact on evaluation, and Clean-Eval \cite{zhu-etal-2024-clean}, which employs neural-based paraphrasing to generate semantically equivalent but surface-level different expressions of potentially contaminated data.
\par
Our research extends these clean evaluation techniques to the visual domain, introducing a novel approach to mitigate data contamination in VLMs through visual data augmentation methods. We collect and present a new clean evaluation benchmark for VLMs and propose various data augmentation techniques, with a novel method called BGR channel swapping emerging as a particularly robust method for clean evaluation. This method demonstrates resistance to exploitation during training, effectively reducing the performance gap between contaminated and uncontaminated models.
\par
This work significantly advances the field of VLMs evaluation, enhancing transparency and reliability in assessing model capabilities. As multi-modal AI continues to evolve, such clean evaluation methods will play a crucial role in ensuring the integrity of model development and deployment. Future research may focus on extending these approaches to other modalities and investigating their potential to improve the robustness and generalization capabilities of VLMs beyond clean evaluation. Our resources will be released upon publication.
\section*{Limitations}
This paper has studied visual data contamination on question answering. Further extending the scope of tasks can enhance the usefulness of the method.
\section*{Ethics Statement}
We honour and support the ARR Code of Ethics. We spot no obvious ethical issues in this paper.
\bibliography{latex/custom}

\begin{thebibliography}{17}
\providecommand{\natexlab}[1]{#1}

\bibitem[{Cao et~al.(2024)Cao, Buchner, Senane, and Yang}]{cao-etal-2024-introducing}
Lele Cao, Valentin Buchner, Zineb Senane, and Fangkai Yang. 2024.
\newblock \href {https://doi.org/10.18653/v1/2024.trustnlp-1.16} {Introducing {G}en{C}eption for multimodal {LLM} benchmarking: You may bypass annotations}.
\newblock In \emph{Proceedings of the 4th Workshop on Trustworthy Natural Language Processing (TrustNLP 2024)}, pages 196--201, Mexico City, Mexico. Association for Computational Linguistics.

\bibitem[{Dong et~al.(2024)Dong, Jiang, Liu, Jin, Gu, Yang, and Li}]{dong-etal-2024-generalization}
Yihong Dong, Xue Jiang, Huanyu Liu, Zhi Jin, Bin Gu, Mengfei Yang, and Ge~Li. 2024.
\newblock \href {https://doi.org/10.18653/v1/2024.findings-acl.716} {Generalization or memorization: Data contamination and trustworthy evaluation for large language models}.
\newblock In \emph{Findings of the Association for Computational Linguistics ACL 2024}, pages 12039--12050, Bangkok, Thailand and virtual meeting. Association for Computational Linguistics.

\bibitem[{Fan et~al.(2024)Fan, Hua, and Li}]{Fan2024nphardeval4v:}
Lizhou Fan, Wenyue Hua, and Xiang Li. 2024.
\newblock \href {https://arxiv.org/abs/2403.01777} {Nphardeval4v: A dynamic reasoning benchmark of multimodal large language models}.
\newblock \url{https://synthical.com/article/8cc2ab7b-0527-4a90-912c-9bab53ecc10a}.
\newblock \emph{Preprint}, arXiv:2403.01777.

\bibitem[{Fan et~al.(2023)Fan, Li, Ma, Lee, Yu, and Hemphill}]{Fan2023ABR}
Lizhou Fan, Lingyao Li, Zihui Ma, Sanggyu Lee, Huizi Yu, and Libby Hemphill. 2023.
\newblock \href {https://api.semanticscholar.org/CorpusID:257952516} {A bibliometric review of large language models research from 2017 to 2023}.
\newblock \emph{ACM Transactions on Intelligent Systems and Technology}.

\bibitem[{Fu et~al.(2023{\natexlab{a}})Fu, Chen, Shen, Qin, Zhang, Lin, Qiu, Lin, Yang, Zheng, Li, Sun, and Ji}]{Fu2023MMEAC}
Chaoyou Fu, Peixian Chen, Yunhang Shen, Yulei Qin, Mengdan Zhang, Xu~Lin, Zhenyu Qiu, Wei Lin, Jinrui Yang, Xiawu Zheng, Ke~Li, Xing Sun, and Rongrong Ji. 2023{\natexlab{a}}.
\newblock \href {https://api.semanticscholar.org/CorpusID:259243928} {Mme: A comprehensive evaluation benchmark for multimodal large language models}.
\newblock \emph{ArXiv}, abs/2306.13394.

\bibitem[{Fu et~al.(2023{\natexlab{b}})Fu, Zhang, Wang, Huang, Zhang, Qiu, Ye, Shen, Zhang, Chen, Zhao, Lin, Jiang, Yin, Gao, Li, Li, and Sun}]{Fu2023ACT}
Chaoyou Fu, Renrui Zhang, Zihan Wang, Yubo Huang, Zhengye Zhang, Longtian Qiu, Gaoxiang Ye, Yunhang Shen, Mengdan Zhang, Peixian Chen, Sirui Zhao, Shaohui Lin, Deqiang Jiang, Di~Yin, Peng Gao, Ke~Li, Hongsheng Li, and Xing Sun. 2023{\natexlab{b}}.
\newblock \href {https://api.semanticscholar.org/CorpusID:266362555} {A challenger to gpt-4v? early explorations of gemini in visual expertise}.
\newblock \emph{ArXiv}, abs/2312.12436.

\bibitem[{Hu et~al.(2021)Hu, Shen, Wallis, Allen{-}Zhu, Li, Wang, and Chen}]{DBLP:journals/corr/abs-2106-09685}
Edward~J. Hu, Yelong Shen, Phillip Wallis, Zeyuan Allen{-}Zhu, Yuanzhi Li, Shean Wang, and Weizhu Chen. 2021.
\newblock \href {https://arxiv.org/abs/2106.09685} {Lora: Low-rank adaptation of large language models}.
\newblock \emph{CoRR}, abs/2106.09685.

\bibitem[{Huang et~al.(2024)Huang, Yan, Li, and Peng}]{app14125068}
Dawei Huang, Chuan Yan, Qing Li, and Xiaojiang Peng. 2024.
\newblock \href {https://doi.org/10.3390/app14125068} {From large language models to large multimodal models: A literature review}.
\newblock \emph{Applied Sciences}, 14(12).

\bibitem[{Jacovi et~al.(2023)Jacovi, Caciularu, Goldman, and Goldberg}]{jacovi-etal-2023-stop}
Alon Jacovi, Avi Caciularu, Omer Goldman, and Yoav Goldberg. 2023.
\newblock \href {https://doi.org/10.18653/v1/2023.emnlp-main.308} {Stop uploading test data in plain text: Practical strategies for mitigating data contamination by evaluation benchmarks}.
\newblock In \emph{Proceedings of the 2023 Conference on Empirical Methods in Natural Language Processing}, pages 5075--5084, Singapore. Association for Computational Linguistics.

\bibitem[{Lin(2004)}]{lin-2004-ROUGE}
Chin-Yew Lin. 2004.
\newblock \href {https://aclanthology.org/W04-1013} {{ROUGE}: A package for automatic evaluation of summaries}.
\newblock In \emph{Text Summarization Branches Out}, pages 74--81, Barcelona, Spain. Association for Computational Linguistics.

\bibitem[{Magar and Schwartz(2022)}]{magar-schwartz-2022-data}
Inbal Magar and Roy Schwartz. 2022.
\newblock \href {https://doi.org/10.18653/v1/2022.acl-short.18} {Data contamination: From memorization to exploitation}.
\newblock In \emph{Proceedings of the 60th Annual Meeting of the Association for Computational Linguistics (Volume 2: Short Papers)}, pages 157--165, Dublin, Ireland. Association for Computational Linguistics.

\bibitem[{Papineni et~al.(2002)Papineni, Roukos, Ward, and Zhu}]{papineni-etal-2002-bleu}
Kishore Papineni, Salim Roukos, Todd Ward, and Wei-Jing Zhu. 2002.
\newblock \href {https://doi.org/10.3115/1073083.1073135} {{B}leu: a method for automatic evaluation of machine translation}.
\newblock In \emph{Proceedings of the 40th Annual Meeting of the Association for Computational Linguistics}, pages 311--318, Philadelphia, Pennsylvania, USA. Association for Computational Linguistics.

\bibitem[{Radford et~al.(2021)Radford, Kim, Hallacy, Ramesh, Goh, Agarwal, Sastry, Askell, Mishkin, Clark et~al.}]{radford2021learning}
Alec Radford, Jong~Wook Kim, Chris Hallacy, Aditya Ramesh, Gabriel Goh, Sandhini Agarwal, Girish Sastry, Amanda Askell, Pamela Mishkin, Jack Clark, et~al. 2021.
\newblock Learning transferable visual models from natural language supervision.
\newblock In \emph{International conference on machine learning}, pages 8748--8763. PMLR.

\bibitem[{Zhang et~al.(2024)Zhang, Sun, Yeats, Ouyang, Kuo, Zhang, Yang, and Li}]{zhang2024min}
Jingyang Zhang, Jingwei Sun, Eric Yeats, Yang Ouyang, Martin Kuo, Jianyi Zhang, Hao Yang, and Hai Li. 2024.
\newblock Min-k\%++: Improved baseline for detecting pre-training data from large language models.
\newblock \emph{arXiv preprint arXiv:2404.02936}.

\bibitem[{Zhao et~al.(2023)Zhao, Zhou, Li, Tang, Wang, Hou, Min, Zhang, Zhang, Dong, Du, Yang, Chen, Chen, Jiang, Ren, Li, Tang, Liu, Liu, Nie, and rong Wen}]{Zhao2023ASO}
Wayne~Xin Zhao, Kun Zhou, Junyi Li, Tianyi Tang, Xiaolei Wang, Yupeng Hou, Yingqian Min, Beichen Zhang, Junjie Zhang, Zican Dong, Yifan Du, Chen Yang, Yushuo Chen, Z.~Chen, Jinhao Jiang, Ruiyang Ren, Yifan Li, Xinyu Tang, Zikang Liu, Peiyu Liu, Jianyun Nie, and Ji~rong Wen. 2023.
\newblock \href {https://api.semanticscholar.org/CorpusID:257900969} {A survey of large language models}.
\newblock \emph{ArXiv}, abs/2303.18223.

\bibitem[{Zhu et~al.(2024{\natexlab{a}})Zhu, Chen, Wang, Gong, Yang, and Xie}]{zhu2024dyval}
Kaijie Zhu, Jiaao Chen, Jindong Wang, Neil~Zhenqiang Gong, Diyi Yang, and Xing Xie. 2024{\natexlab{a}}.
\newblock \href {https://openreview.net/forum?id=gjfOL9z5Xr} {Dyval: Dynamic evaluation of large language models for reasoning tasks}.
\newblock In \emph{The Twelfth International Conference on Learning Representations}.

\bibitem[{Zhu et~al.(2024{\natexlab{b}})Zhu, Hao, He, Song, Yueyang, Zhang, Hu, Wei, Wang, and Lu}]{zhu-etal-2024-clean}
Wenhong Zhu, Hongkun Hao, Zhiwei He, Yun-Ze Song, Jiao Yueyang, Yumeng Zhang, Hanxu Hu, Yiran Wei, Rui Wang, and Hongyuan Lu. 2024{\natexlab{b}}.
\newblock \href {https://doi.org/10.18653/v1/2024.findings-naacl.53} {{CLEAN}{--}{EVAL}: Clean evaluation on contaminated large language models}.
\newblock In \emph{Findings of the Association for Computational Linguistics: NAACL 2024}, pages 835--847, Mexico City, Mexico. Association for Computational Linguistics.

\end{thebibliography}

\end{document}